\documentclass[runningheads]{llncs}
\usepackage[T1]{fontenc}

\usepackage{graphicx}
\usepackage{verbatim}
\usepackage{amsmath}
\usepackage{caption}
\usepackage{hyperref}
\usepackage{xcolor}
\usepackage{booktabs, amssymb, siunitx, xcolor, colortbl}

\hypersetup{
  colorlinks=true,
  urlcolor=blue,
  linkcolor=black,
  citecolor=black
}

\begin{document}

\title{DiffUS: Differentiable Ultrasound Rendering from Volumetric Imaging}
\author{Noe Bertramo \and Gabriel Duguey \and Vivek Gopalakrishnan}
\authorrunning{N. Bertramo \textit{et al.}}
\institute{Massachusetts Institute of Technology, Cambridge, MA, USA \\
\email{\{noe\_bert,gduguey,vivekg\}@mit.edu}}

\maketitle
\begin{abstract}
Intraoperative ultrasound imaging provides real-time guidance during numerous surgical procedures, but its interpretation is complicated by noise, artifacts, and poor alignment with high-resolution preoperative MRI/CT scans. To bridge the gap between preoperative planning and intraoperative guidance, we present DiffUS, a physics-based, differentiable ultrasound renderer that synthesizes realistic B-mode images from volumetric imaging. DiffUS first converts MRI 3D scans into acoustic impedance volumes using a machine learning approach. Next, we simulate ultrasound beam propagation using ray tracing with coupled reflection-transmission equations. DiffUS formulates wave propagation as a sparse linear system that captures multiple internal reflections. Finally, we reconstruct B-mode images via depth-resolved echo extraction across fan-shaped acquisition geometry, incorporating realistic artifacts including speckle noise and depth-dependent degradation. DiffUS is entirely implemented as differentiable tensor operations in PyTorch, enabling gradient-based optimization for downstream applications such as slice-to-volume registration and volumetric reconstruction. 
Evaluation on the ReMIND dataset demonstrates DiffUS's ability to generate anatomically accurate ultrasound images from brain MRI data. Our implementation is available at \href{https://github.com/gduguey/DiffUS}{https://github.com/gduguey/DiffUS}.
\end{abstract}

\section{Introduction}
Ultrasound (US) imaging is valuable for intraoperative guidance given its speed, safety, and portability. However, US images are plagued by noise, artifacts, and poor contrast, complicating anatomical interpretation in delicate tasks such as tumor resection~\cite{MACHADO2019116094}. Furthermore, the difficulty in aligning intraoperative US and high-resolution preoperative MRI or CT scans forces surgeons to perform a complex mental mapping across modalities with disparate visual characteristics. This disconnect can make it difficult to accurately locate critical anatomical structures, identify residual pathology, and adjust surgical plans in real time. Generating realistic ultrasound images from preoperative volumetric scans could help overcome these challenges by improving spatial alignment between modalities and supporting more effective surgical guidance.

To this end, we propose DiffUS, a physics-based, differentiable ultrasound renderer that models acoustic wave propagation through coupled reflection-transmission equations. Unlike previous ultrasound simulation methods that rely on random sampling~\cite{duelmer2025ultrarayfullpathraytracing}, our method employs deterministic ray propagation with sequential processing to capture temporal echo characteristics. Furthermore, our approach formulates wave propagation as a stable linear system solution, enabling efficient computation of multiple internal reflections that is faster than existing approaches~\cite{treeby2020nonlinear}. The differentiable nature of our renderer facilitates gradient-based optimization for various downstream applications, including registration and reconstruction tasks.

Our work contributes to the growing field of physics-based medical imaging simulation~\cite{billot2023synthseg,gao2023synthetic,gopalakrishnan2022fast} by providing an interpretable, computationally efficient framework for ultrasound synthesis. DiffUS rapidly generates realistic ultrasound images towards the ultimate goal of supporting improved surgical outcomes through enhanced multimodal imaging integration.

\section{Method}
DiffUS is a differentiable, physics-based US renderer that synthesizes realistic B-mode images by modeling acoustic wave propagation through volumetric MRI or CT scans. 

In the following sections, we detail the fundamental physics principles underlying our simulator (\S\ref{sec:physics}), our method for mapping CT/MRI intensities to acoustic impedances (\S\ref{sec:impedance}), our formulation of wave propagation as a sparse linear system (\S\ref{sec:forward_model}), and the final image formation model (\S\ref{sec:image_synthesis}).

\subsection{Physical Principles of Ultrasound Imaging}
\label{sec:physics}
US imaging captures the propagation of acoustic waves through biological tissues at frequencies ranging from 2-15 MHz. The physics of US is governed by the wave equation, with propagation characteristics determined by acoustic impedance $Z = \rho c$, where $\rho$ denotes tissue density and $c$ the speed of sound. At tissue interfaces, wave behavior follows the reflection and transmission coefficients:
\begin{equation}
R_{1 \rightarrow 2} = \left|\frac{Z_2 - Z_1}{Z_1 + Z_2}\right|, \quad T_{1 \rightarrow 2} = \frac{2Z_2}{Z_2 + Z_1}
\end{equation}
where $R$ and $T$ represent reflection and transmission coefficients between media with impedances $Z_1$ and $Z_2$.
Clinical ultrasound systems operate under several simplifying assumptions \cite{aldrich2007basic,huang2007imaging} that we incorporate in DiffUS: (1) acoustic waves propagate along straight-line paths, (2) the speed of sound is constant across soft tissues ($c = 1540$ m/s), (3) acquisitions represent a single pulse-echo, (4) attenuation is uniform across media, and (5) the contribution of side-lobes is negligible. Under these conditions, the reflecting structure depth is determined by the time-of-flight measurement $d = \frac{c}{2}(t_2 - t_1)$, where $t_1$ and $t_2$ represent pulse emission and echo reception times, respectively.

\subsection{Generating Acoustic Impedance Volumes}
\label{sec:impedance}
CT and MRI volumes do not directly encode impedance. Instead, the impedance of every voxel must be inferred through indirect relationships to underlying physical or biological properties. Therefore, to model acoustic wave propagation, we convert volumetric medical images into three-dimensional acoustic impedance maps $Z(x,y,z)$, which  determine the intensity of both reflected and transmitted signals at tissue interfaces. 

\paragraph{CT-based impedance mapping.}
Voxels in CT are encoded using Hounsfield units (HU), calibrated quantities that reflect the local X-ray attenuation coefficient relative to water~\cite{levine2018preliminary}. Because X-ray attenuation in biological tissues is dominated by Compton scattering at clinical energies (70–140~kVp), and Compton scattering scales approximately with electron density, HU values provide a strong proxy for physical density~\cite{schneider1996calibration}. Therefore, we use an existing calibration method that fits a piecewise linear relationship between HU and density across typical biological tissues~\cite{schneider1996calibration,wein2008automatic}. From this density map $\rho(x,y,z)$, we compute acoustic impedance as $Z(x,y,z) = \rho(x,y,z) \cdot c(x,y,z)$, where $c$ is estimated directly from HU using previously reported regression fits~\cite{webb2018measurements}. This gives us a voxel-wise impedance map grounded in physical tissue characteristics.

\begin{figure}[t!]
    \centering
    \includegraphics[width=0.7\textwidth]{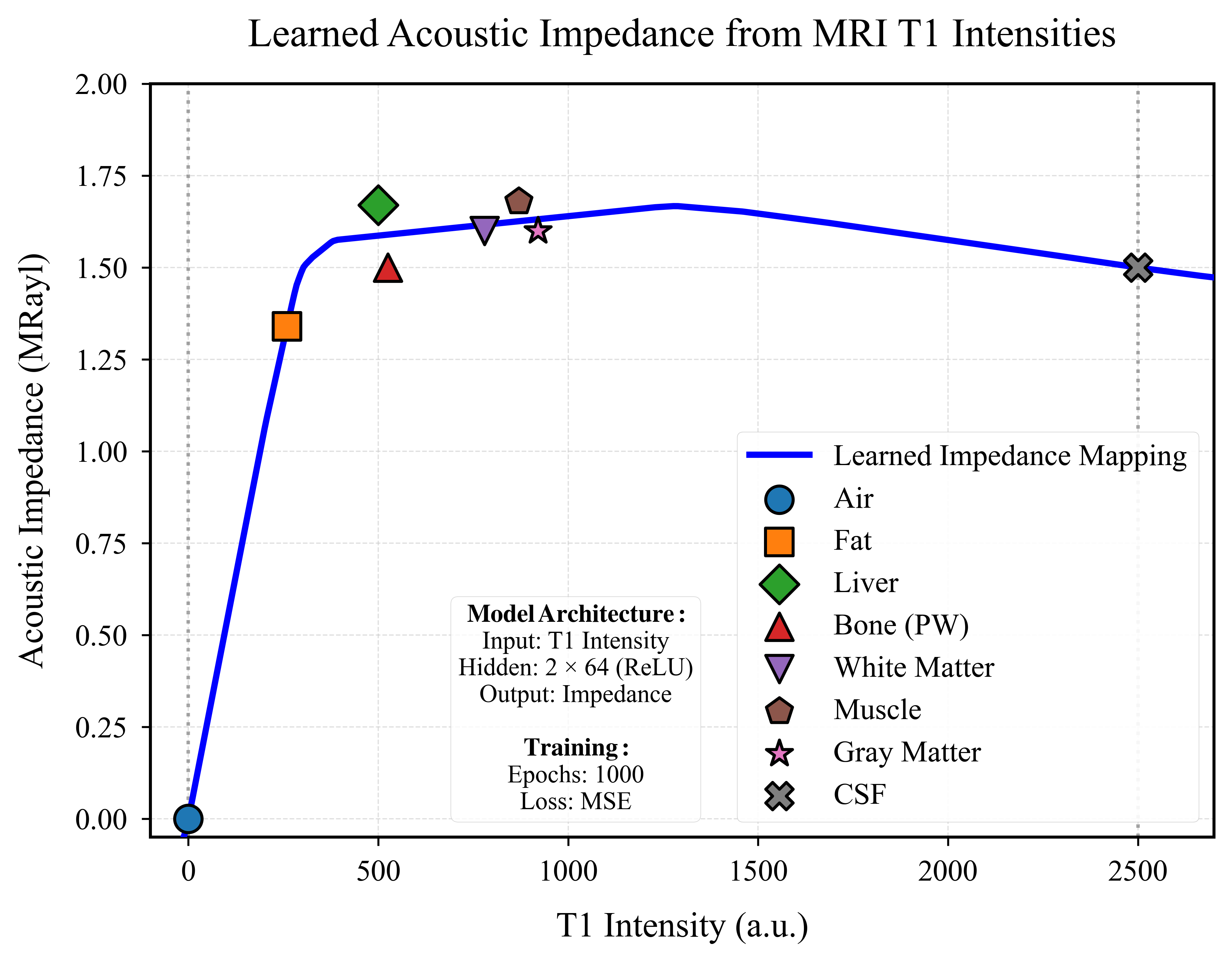}
    \caption{\textbf{MLP-based acoustic impedance estimation from MRI T1 intensities.} (a) The learned continuous mapping (blue curve) transforms T1 intensities to acoustic impedance values (MRayl). (b) Tissue-specific reference points (markers) used for supervised training.}
    \label{fig:mlp_mapping}
\end{figure}

\paragraph{MRI-based impedance mapping.}
Unlike CT, MRI intensities do not correspond directly to physical quantities such as density or acoustic impedance. Instead, they reflect a complex combination of tissue-specific properties, including water content, fat concentration, and relaxation times, modulated by the chosen acquisition sequence. This makes impedance reconstruction from MRI fundamentally ill-posed~\cite{carre2020standardization}. To approximate impedance, we therefore adopt a supervised learning approach, fitting a multi-layer perceptron (MLP) to paired data consisting of MRI intensities and reference impedance values from literature~\cite{chan2011basics,chen2017measurement,drew2025t1,joarder2020uplifted}. While this provides a continuous and flexible estimate of $Z(x,y,z)$ across the volume, it is a heuristic approximation that lacks the physical interpretability of CT-derived methods. The training data and learned mapping are illustrated in Figure~\ref{fig:mlp_mapping}.

\subsection{Wave Propagation Model}
\label{sec:forward_model}
\begin{figure}[t]
    \centering
    \includegraphics[width=\linewidth]{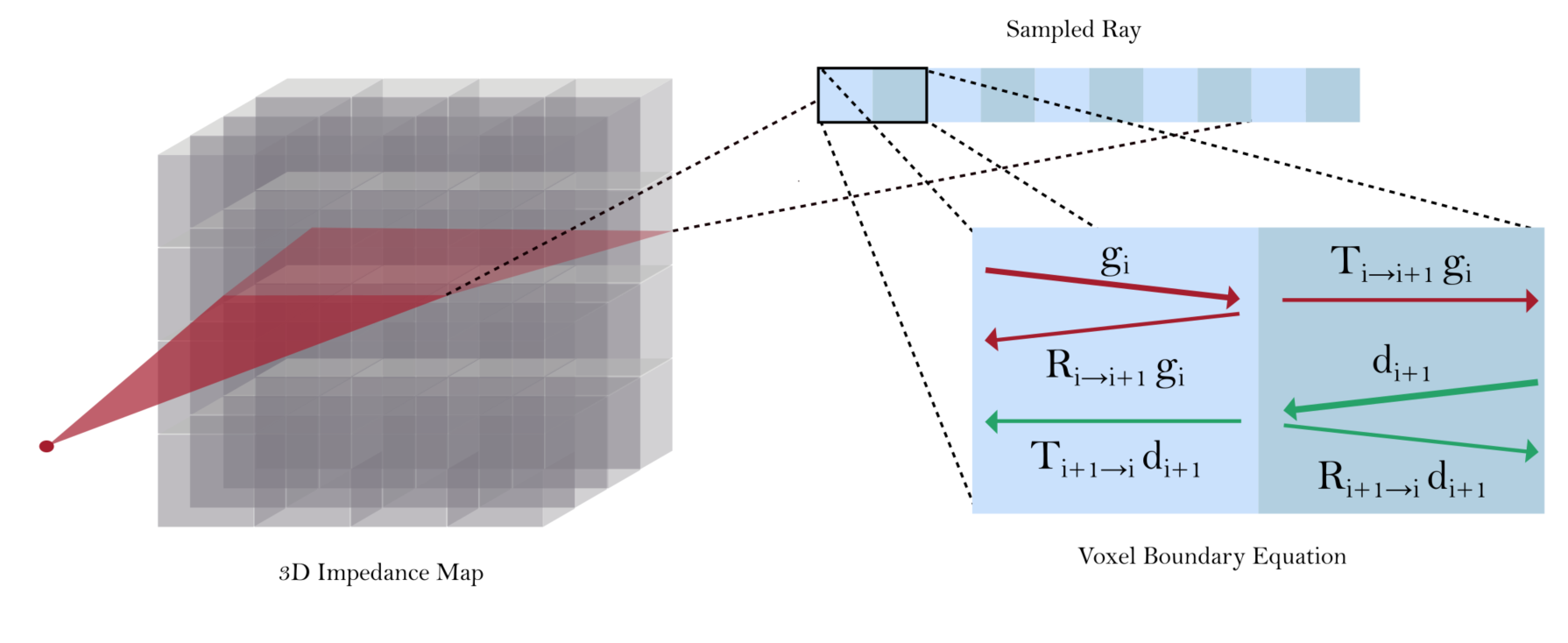}
    \caption{\textbf{Modelling wave propagation in DiffUS.} Coupled recursion relations for forward ($g_i$) and backward ($d_i$) wave amplitudes across an interface.}
    \label{fig:ray_recursion}
\end{figure}

We model the propagation of the ultrasound beam using a ray tracing approach where each acoustic beam traverses a linear sequence of voxels. At each voxel boundary, wave amplitudes undergo reflection and transmission according to local impedance contrasts. We define forward-traveling amplitude $g_i$ and backward-traveling amplitude $d_i$ at each interface $i,i+1$, which evolve according to coupled recurrence relations:
\begin{equation}
g_{i+1} = T_{i \rightarrow i+1} \cdot g_i + R_{i+1\rightarrow i} \cdot d_{i+1} \
\quad \quad d_i = R_{i\rightarrow i+1} \cdot g_i + T_{i+1 \rightarrow i} \cdot d_{i+1}
\end{equation}
These equations capture multiple internal reflections and transmissions throughout the layered medium, providing physically accurate wave amplitude evolution. The coupling logic between incident 
and reflected waves across a single boundary is illustrated in Figure~\ref{fig:ray_recursion}.

We solve these coupled wave equations by formulating a sparse linear system $\mathbf{A} \mathbf{x} = \mathbf{b}$ for $N$ voxels. We arrange amplitude unknowns as
\begin{equation}
    \mathbf{x} = [g_0, d_0, g_1, d_1, \ldots, g_N, d_N]^T   
\end{equation}
where boundary conditions impose unit pulse injection ($g_0 = 1$) and far-field termination ($d_N = 0$). These boundary conditions are enforced through the right-hand side vector $\mathbf{b}$, which is zero everywhere except at the first entry ($b_0 = 1$) to inject a unit-amplitude forward-traveling wave at the transducer. This choice of $\mathbf{b}$, in combination with the identity rows $A_{0,0} = 1$ and $A_{2N+1,2N+1} = 1$, ensures that the boundary constraints $g_0 = 1$ and $d_N = 0$ are directly imposed within the linear system. The remaining entries of the matrix $\mathbf{A} = (A_{r,c})$ enforce the recurrence relations for wave amplitudes at each interface:
\small
\begin{equation}
    A_{r,c} = \begin{cases}
    1 & r=0,\ c=0 \\
    1 & r=2N+1,\ c=2N+1 \\
    -R_{i\rightarrow i+1} & r=2i+1,\ c=2i \\
    1 & r=2i+1,\ c=2i+1 \\
    -T_{i+1\rightarrow i} & r=2i+1,\ c=2i+3 \\
    -T_{i\rightarrow i+1} & r=2i+2,\ c=2i \\
    1 & r=2i+2,\ c=2i+2 \\
    -R_{i+1\rightarrow i} & r=2i+2,\ c=2i+3 \\
    0 & \text{otherwise}
    \end{cases}
\end{equation}
\normalsize
for $i=0,\ldots,N-1$ and $r,c=0,\ldots,2N+1$.
The matrix $\mathbf{A}$ is sparse, with each row involving only a small number of neighboring amplitude variables due to the strictly local nature of wave interactions. This results in a fixed-bandwidth structure with at most three non-zero entries per row, leading to $O(N)$ non-zero elements for $N$ voxels. While our current implementation uses a dense solver for simplicity, the sparsity of $\mathbf{A}$ permits more efficient factorization, which could be exploited in future work using banded linear solvers to accelerate computation.

\subsection{Image Formation Model and Artifact Modeling}
\label{sec:image_synthesis}
Ultrasound image synthesis follows a depth-resolved approach. For each beam direction, we progressively solve wave systems at increasing voxel depths \(k=1,\dots,N\), measuring the returning echo amplitude $d_0^{(k)}$ at the transducer location at each step. This generates a complete depth profile $\{d_0^{(1)}, \ldots, d_0^{(N)}\}$ representing the time-sequence of echoes received by the transducer. B-mode images are constructed by repeating this process across multiple ray directions from the virtual transducer position, creating a fan-shaped acquisition pattern that replicates clinical scanning geometry. Pixel intensities correspond to echo amplitude first differences, while spatial coordinates map to beam angle and depth. 

To enhance clinical realism, DiffUS applies two optional post-processing stages that reproduce characteristic ultrasound artifacts. First, speckle can be injected as multiplicative noise with depth-dependent amplitude through a radially oriented term whose strength follows a power-law with depth and a granular term that imitates tissue heterogeneity. Both strength and granularity intensify as the signal-to-noise ratio declines in deeper regions. Second, depth-related blurring is imposed by convolving the image with a Gaussian kernel whose width grows linearly with depth, capturing the lateral resolution loss caused by beam divergence and energy attenuation. The artifact parameters can be adjusted to match different scanners, transducers, or imaging settings.

\section{Results}
\label{sec:results}

\subsection{Dataset and Experimental Setup}
We evaluate the simulation fidelity of DiffUS using the ReMIND dataset~\cite{juvekar2023brain}, a collection of rigidly aligned pairs of preoperative T1-weighted MRI and intraoperative 3D ultrasound (iUS) volumes from brain tumor resections. Specifically, we convert T1 intensities to acoustic impedance via our trained MLP (Section~\ref{sec:impedance}) and render synthetic ultrasounds using a fan-beam acquisition with 256 rays and 100-200 depth samples per ray.

\subsection{Qualitative Assessment of Anatomical Fidelity}

\begin{figure}[b!]
\centering
\includegraphics[width=\linewidth]{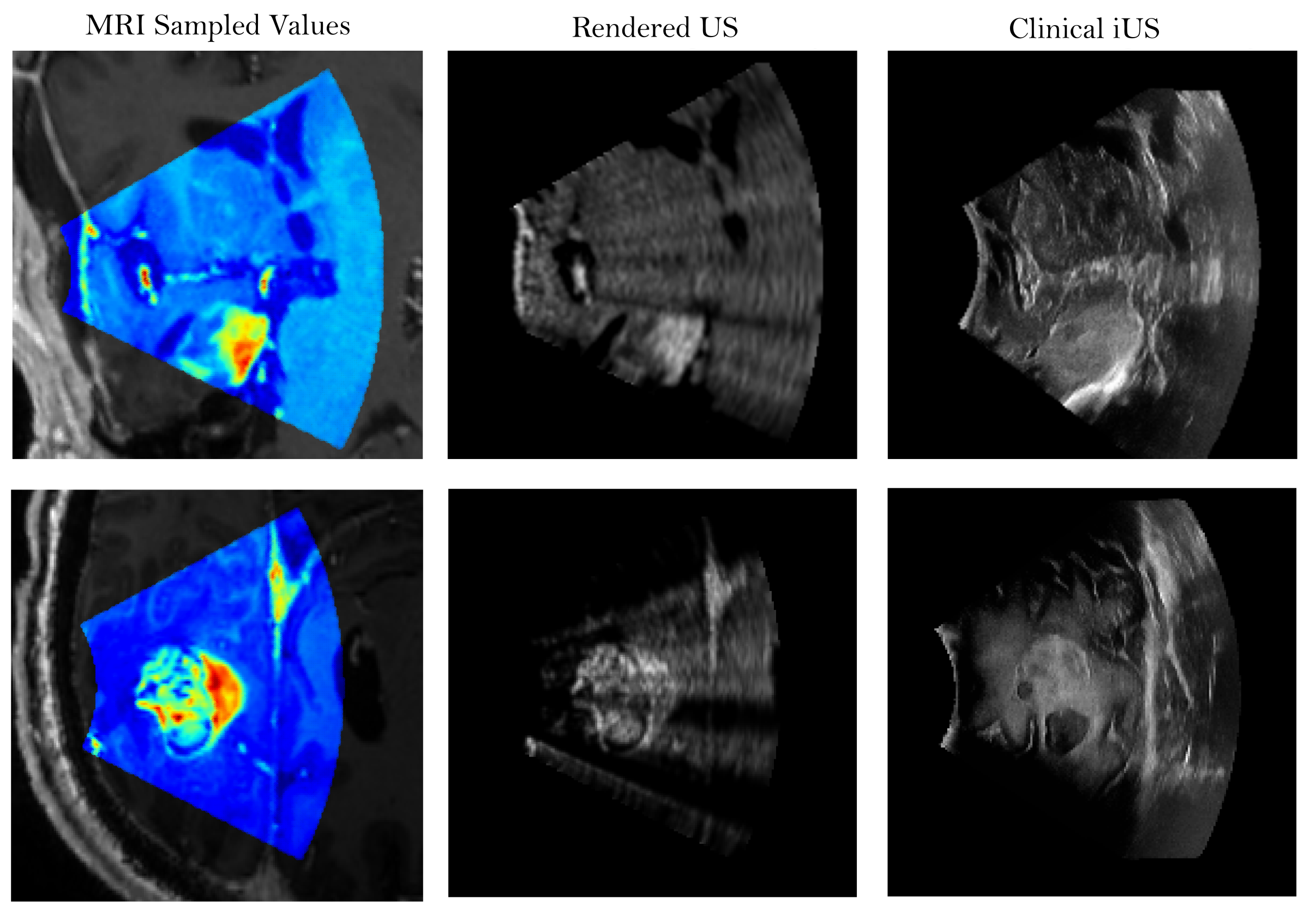}
\caption{\textbf{Qualitative evaluation on brain MRI data from the ReMIND dataset.} (Left) Impedance ray sampling pattern overlaid on preoperative MRI. (Center) Ultrasound image generated by DiffUS. (Right) Corresponding ground truth iUS acquisition. Note that non-rigid deformation between the preoperative MRI volume and iUS images due to brain shift is not accounted for.}
\label{fig:MRI}
\end{figure}

Figure~\ref{fig:MRI} compares rendered ultrasound images generated by DiffUS (center) against corresponding tracked iUS acquisitions (right) for two representative cases. The left column depicts the acoustic impedance values predicted by our MLP for each voxel superimposed onto the T1-weighted MRI data, illustrating the transducer geometry and acoustic beam propagation trajectories.

These images demonstrate that DiffUS successfully recapitulates the geometry of gross neuroanatomical features and their relative acoustic impedance values. Most strikingly, the ventricles, which appear as hypointense regions in the MRI, consistently appear as hypoechoic areas on the rendered ultrasound images. Additionally, the ventricles in the rendered ultrasounds appear radially warped relative to the MRI volume, validating the transducer geometry implemented in DiffUS. The second row of Figure~\ref{fig:MRI} also demonstrates that the sulcus at the anterior portion of the brain is properly rendered by DiffUS. The sulcus also appears hyperintense relative to the surrounding tissue, albeit with less pronounced contrast compared to the clinical iUS. These similarities in geometry and intensity demonstrate the fidelity of our MRI-to-impedance network and demonstrate that DiffUS accurately maps anatomical information encoded in preoperative MRI volumes to rendered ultrasounds.

\subsection{Ablation Study}
We quantify the contribution of each component of our pipeline through a comprehensive ablation study. We first consider geometrically resampling the MRI to match an iUS acquisition pattern. This baseline mimics the mental alignment a surgeon might attempt between MRI and iUS images. Next, we progressively ablate (i) Impedance Mapping (\S\ref{sec:impedance}), (ii) Ray Propagation (\S\ref{sec:forward_model}), and (iii) Artifact Modeling (\S\ref{sec:image_synthesis}). We compare the resulting rendered and real iUS images using Mean Squared Error (MSE), Structural Similarity Index (SSIM), Normalized Cross-Correlation (NCC), and Mean Absolute Error (MAE).

\begin{table}[t]
\centering
\caption{Ablation study on reconstruction metrics (mean~(std), $\uparrow$/$\downarrow$ show direction of improvement). The two highlighted rows show the full pipeline, with and without optional artifacts that improve visualization but hurt metrics.}
\vspace{0.4em}
\setlength{\tabcolsep}{2pt}
\label{tab:ablation}
\resizebox{\textwidth}{!}{%
\begin{tabular}{cccccccc}
\toprule
Sampling & Mapping & Propagating & Artifacts &
{MSE ($\downarrow$)} & {SSIM ($\uparrow$)} & {NCC ($\uparrow$)} & {MAE ($\downarrow$)} \\
\midrule
\checkmark &        &        &              & 0.02(0.01) & 0.65(0.01) & 0.85(0.05) & 0.07(0.02) \\
\checkmark & \checkmark &        &              & 0.15(0.07) & 0.61(0.03) & 0.86(0.04) & 0.22(0.06) \\
\rowcolor{gray!20}
\checkmark & \checkmark & \checkmark &         & 0.05(0.03) & \textbf{0.76(0.06)} & \textbf{0.93(0.04)} & 0.11(0.03) \\
\rowcolor{gray!20}
\checkmark & \checkmark & \checkmark & \checkmark & 0.03(0.01) & 0.62(0.04) & 0.85(0.09) & 0.08(0.03) \\
\checkmark &        & \checkmark &              & 0.03(0.02) & 0.75(0.06) & 0.92(0.05) & 0.08(0.02) \\
\checkmark &        & \checkmark & \checkmark   & \textbf{0.02(0.01)} & 0.63(0.02) & 0.81(0.05) & \textbf{0.07(0.01)} \\
\bottomrule
\end{tabular}
}
\vspace{-1em}
\end{table}

\begin{figure}[h]
    \vspace{-1em}
    \centering
    \includegraphics[width=\linewidth]{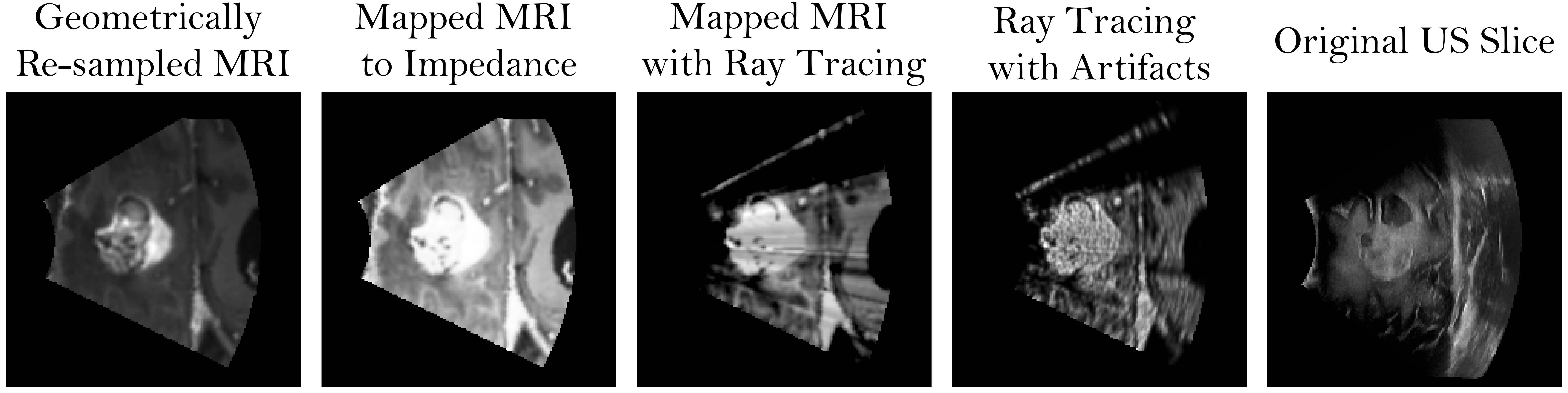}
    \caption{Detailed images for each ablation configuration and the reference iUS.}
    \label{fig:ablation}
\end{figure}

Table~\ref{tab:ablation} shows improvements in similarity when progressively adding physics-based modules beyond the geometric baseline. While artifact modeling improves visual realism, it can slightly decrease pixel-wise similarity due to added texture not present in the reference ultrasound. A general limitation of these metrics is their reliance on strict pixel alignment between predicted and reference images. This assumption can be violated in practice due to factors such as tissue deformation or brain shift during surgery. To reduce rigid misalignment errors, we apply a phase cross-correlation registration step before metric computation. Nonetheless, perfect alignment cannot be guaranteed and should be considered when interpreting these values. Figure~\ref{fig:ablation} illustrates qualitative results for the main ablation configurations compared with the post-operative ultrasound slice.

\subsection{Computational Performance Analysis}

We evaluated the computational efficiency of DiffUS on an AMD EPYC 9474F @ 3.6 GHz CPU and an NVIDIA L40S 48GB GPU. For the standard imaging parameters used in our brain experiments (256 acoustic rays and 100 to 200 depth samples per ray), a single 2D fan-shaped slice renders in around one to two seconds for typical depths, as detailed in Tables~\ref{tab:rendering_times_cpu} and~\ref{tab:rendering_times_gpu}. Here, depth corresponds to the number of voxels sampled in each ray. 

\begin{figure}[h]
\centering
\begin{minipage}[t]{0.48\textwidth}
\centering
\captionof{table}{Rendering time (seconds) for varying depth samples and ray counts over 100 iterations on CPU.}
\label{tab:rendering_times_cpu}
\begin{tabular}{@{}c@{~}|@{~}ccc@{}}
\toprule
\textbf{Samples} & \textbf{64 rays} & \textbf{128 rays} & \textbf{256 rays} \\ \midrule
32& 0.04 s& 0.05 s& 0.06 s\\
70& 0.21 s& 0.27 s& 0.40 s\\
100& 0.47 s& 0.69 s& 1.15 s\\
200 & 3.20 s& 6.00 s& 10.50 s\\ \bottomrule
\end{tabular}
\end{minipage}%
\hfill
\begin{minipage}[t]{0.48\textwidth}
\centering
\captionof{table}{Rendering time (seconds) for varying depth samples and ray counts over 100 iterations on GPU.}
\label{tab:rendering_times_gpu}
\begin{tabular}{@{}c@{~}|@{~}ccc@{}}
\toprule
\textbf{Samples} & \textbf{64 rays} & \textbf{128 rays} & \textbf{256 rays} \\ \midrule
32 & 0.04 s & 0.04 s & 0.04 s \\
70 & 0.20 s & 0.21 s & 0.21 s \\
100 & 0.41 s & 0.42 s & 0.43 s \\
200 & 1.67 s & 1.70 s & 1.84 s \\ \bottomrule
\end{tabular}
\end{minipage}
\end{figure}

\section{Conclusion}
We present DiffUS, a fast, differentiable, and physics-based framework for differentiable ultrasound rendering from preoperative MRI and CT volumes. DiffUS uses a novel formulation of wave propagation as a sparse system of equations to achieve efficient and accurate image synthesis, with downstream applications to many real-time diagnostic and surgical settings. Future application areas include the development of patient-specific networks for ultrasound-to-MRI registration~\cite{gopalakrishnan2025rapid,gopalakrishnan2024intraoperative}, or slice-to-volume reconstruction approaches for fetal ultrasound~\cite{gholipour2010robust,xu2023nesvor,young2024fully}. To inspire such research in this direction, we have released our open-source implementation at \href{https://github.com/gduguey/DiffUS}{https://github.com/gduguey/DiffUS}.

\begin{credits}
\subsubsection{\ackname} We thank Polina Golland, Reuben Dorent, Sandy Wells and Karimi Davood for their valuable insights on ultrasound imaging, MRI data, and questions related to realignment.

\subsubsection{\discintname}
This work was not supported by a specific grant and the authors have no competing interests to declare that are relevant to the content of this article.
\end{credits}
%
%
%
\bibliographystyle{splncs04}
\bibliography{mybibliography}

\end{document}